\documentclass[11pt]{article}
\usepackage[preprint]{neurips_2024}
\usepackage{amsmath,amssymb}
\usepackage{booktabs}
\usepackage{graphicx}
\usepackage{hyperref}
\usepackage{xcolor}

\title{EML-CD: Causal Mechanism Recovery\\via EML Symbolic Trees in Structure Learning}

\author{
  Sota Asanuma \\
  SoftBank Corp.
}
\date{June 4, 2026}

\begin{document}
\maketitle

\begin{abstract}
Neural network (NN)-based nonlinear causal discovery methods recover DAG structure but leave each causal mechanism as a black box.
\citet{waxman2024dagmadce} argued that extracting causal mechanisms from NN weights is ill-posed.
We propose EML-CD, a framework that integrates the EML operator~\citep{odrzywolk2026eml}---capable of composing elementary functions from a single binary operator---into causal structure learning, with interpretable mechanism recovery as the primary objective.
EML-CD represents each edge mechanism as a gated EML binary tree and automatically discovers closed-form causal equations.
Analytical Jacobians can be directly computed from the output equations, enabling quantitative understanding of causal effects.
On real data (Sachs protein signaling, $d{=}11$), EML-CD achieves SHD=$11.2 \pm 0.4$ (5-seed mean; baselines are single deterministic runs), on par with PC/GES~(11) within seed variance and below CAM~(12), while attaching closed-form equations to each detected edge (precision 0.756, recall 0.365).
In a controlled bivariate test with known mechanisms, EML-CD recovers 10 of 11 elementary function families faithfully (held-out shape correlation $\geq 0.96$; only high-frequency sine is partial).
On a symbolic synthetic benchmark, EML-CD attains a substantially lower and more stable held-out mechanism f-MSE than a fixed SINDy dictionary (mean 3.67 vs.\ 7644, the latter inflated by catastrophic extrapolation on one seed), although its structure recovery (SHD 14.0) only matches the dictionary and stays below specialized optimizers; on the Causal Chambers light-tunnel subset, a depth-2 model improves F1 over linear OLS-BIC (0.444 vs.\ 0.273).
\end{abstract}

\section{Introduction}

Causal discovery from observational data has made remarkable progress in DAG recovery.
Continuous optimization methods---NOTEARS~\citep{zheng2018notears}, GOLEM~\citep{ng2020golem}, DAGMA~\citep{bello2022dagma}---achieve near-perfect structure recovery on linear benchmarks, and NN additive noise models (NN-ANMs) such as GraNDAG~\citep{lachapelle2020grandag} and NOTEARS-MLP~\citep{zheng2020notears_mlp} extend this to nonlinear mechanisms.
Yet none of these methods can answer \textbf{what functional form connects each cause to its effect}.

If causal mechanisms were available in closed form, one could quantitatively predict intervention effects and analyze causal effect functions. For instance, a closed-form equation for Raf $\to$ Mek in protein signaling would directly predict how a Raf inhibitor affects Mek levels.

\citet{waxman2024dagmadce} argued that extracting causal mechanisms from NN weights is ill-posed (Lemmas~1, 2).
Post-hoc explanation methods such as SHAP~\citep{lundberg2017shap} and LIME~\citep{ribeiro2016lime} provide only local approximations.
CAM~\citep{buhlmann2014cam} captures nonlinearity via splines, but piecewise polynomials cannot be extracted as closed-form symbolic equations.
BF-BIC~\citep{ramsey2025bfbic} uses truncated basis-function (Legendre) scores for DAG scoring but does not estimate functional forms.
Closed-form mechanisms are recovered jointly with structure only in restricted regimes---linear (e.g., LiNGAM~\citep{shimizu2006lingam}, NOTEARS~\citep{zheng2018notears}) or additive-spline (CAM~\citep{buhlmann2014cam})---while symbolic mechanism extraction on trained models is typically performed post-hoc on a fixed graph~\citep{cranmer2020symbolic}.
To our knowledge, \textbf{no existing method jointly recovers the causal structure and arbitrary nonlinear closed-form mechanisms within a single framework}.

We address this gap by integrating the EML operator $\mathrm{eml}(x, y) = \exp(x) - \ln(y)$~\citep{odrzywolk2026eml} into causal structure learning. The EML operator composes all elementary functions from a single binary operator and constant~1, and a gated binary tree specifies each edge mechanism in finite parameters, learnable by gradient-based optimization.

\textbf{Contributions.}
(1)~We propose EML-CD, a framework that integrates EML symbolic trees into causal structure learning, attaching closed-form equations and analytical Jacobians to each edge.
(2)~On real data (Sachs, SHD=$11.2 \pm 0.4$, 5-seed mean), we show that this interpretability is added while matching PC/GES~(11) within seed variance and staying below CAM~(12).
(3)~In a controlled bivariate test with known mechanisms, EML trees recover 10 of 11 elementary function families faithfully (shape correlation $\geq 0.96$); on a symbolic synthetic benchmark and a Causal Chambers subset we further probe recovery under structure learning and on a real physical system.
(4)~Through concrete examples of detected edges (nonlinear EML compositions and automatic degeneration to linearity), we show that closed-form equations and analytical causal effect functions provide symbolic interpretability unavailable from OLS or splines.

\section{Background and Problem Setting}

To realize the above contributions, we first formalize the theoretical foundations and the ill-posedness problem to be addressed.

\subsection{Additive Noise Models and Ill-Posedness}

We adopt the additive noise model (ANM)~\citep{hoyer2009anm,peters2014anm} and assume causal sufficiency (no unobserved confounders):
$X_j = \sum_{i \in \mathrm{pa}(j)} f_{j,i}(X_i) + \varepsilon_j$.
\citet{buhlmann2014cam} showed that if all $f_{j,i}$ are nonlinear (non-affine), the DAG is identifiable under regularity conditions.

NN-based methods (GraNDAG, NOTEARS-MLP) represent $f_{j,i}$ as neural networks, but as \citet{waxman2024dagmadce} demonstrated, mechanism recovery from NN weights is not identifiable from the learned weights alone.
\textbf{To circumvent this ill-posedness, the function class must be restricted to an interpretable, finite-dimensional space.}
EML-CD achieves this restriction via EML symbolic trees.

\subsection{EML Operator: Symbolic Function Representation as Solution}

\citet{odrzywolk2026eml} showed that the grammar $S \to 1 \mid \mathrm{eml}(S, S)$ is functionally complete for the elementary functions: every elementary function admits an exact EML-tree representation.
Simple functions appear at shallow depth---e.g.\ $\exp(x) = \mathrm{eml}(x, 1)$ (using $\ln 1 = 0$)---whereas others (e.g.\ $\ln$, multiplication, trigonometric functions) require substantially deeper trees~\citep{odrzywolk2026eml}.
EML-CD therefore bounds the tree depth to a small $D$ (\S3.1), trading exact universality for an interpretable, finite-dimensional function class.

EML trees are suitable for mechanism recovery for three reasons:
(i)~a depth-bounded EML tree spans a finite-dimensional, interpretable function class, restricting the mechanism search away from the ill-posed NN weight space;
(ii)~depth gates automatically control function complexity, suppressing overfitting;
(iii)~closed-form equations and analytical Jacobians are directly obtainable from the tree structure.

\section{Proposed Method}

EML-CD (EML-based Causal Discovery) is a causal structure learning method using EML trees as function representation.

\subsection{EML Tree Representation of Causal Mechanisms}

To circumvent the NN ill-posedness described in \S2, each edge mechanism is represented as a depth-$D$ complete binary tree:
\begin{equation}
  X_j = \sum_{i \in \mathrm{pa}(j)} s_{j,i} \cdot T_{j,i}(X_i) + \varepsilon_j, \quad
  \mathrm{node}(l, r) = \sigma(\gamma) \cdot \mathrm{eml}(l, r) + (1 - \sigma(\gamma)) \cdot l
\end{equation}
where $s_{j,i} = \mathrm{softplus}(\alpha_{j,i})$ is the output scale and each leaf is $v_\ell = a_\ell x + b_\ell$.
$\sigma(\gamma)$ is a depth gate: $\gamma \ll 0$ bypasses the node (tree degenerates to linear), $\gamma \gg 0$ activates EML.
This depth gate enables data-driven complexity selection from linear to nonlinear (\S2.2 property~(ii)).
To make the recovered equation \emph{faithful} to the model, we evaluate the gate as $\sigma(\gamma/\tau_g)$ and \textbf{anneal} the gate temperature $\tau_g$ from $1$ to $0.02$ over training (a deterministic-annealing schedule on the relaxation sharpness, not stochastic search; $\tau_g$ is distinct from the R$^2$ edge threshold $\tau$ of \S3.2): early training is soft and gradients flow freely, while at convergence the gates are (near-)binary, so the hard-thresholded closed-form readout equals the soft-gated function it was trained as. Annealing from the start is essential---hard-thresholding a soft-trained tree post hoc destroys the fit (e.g.\ a representative edge drops from R$^2{=}0.67$ to $-0.94$), whereas annealed training reaches comparable hard-gate R$^2$.
At depth $D{=}2$, each edge has 12 parameters ($4\text{ leaves}\times 2 + 3\text{ gates} + 1\text{ scale}$); at $D{=}3$, 24 parameters ($8\text{ leaves}\times 2 + 7\text{ gates} + 1\text{ scale}$).

For numerical stability the EML operator clips its $\exp$ input to $[-2,2]$, floors the $\ln$ argument at $0.5$, and clips node outputs to $[-10,10]$; a custom VJP additionally clips each node's gradient to $[-10,10]$.
These bounds are load-bearing rather than cosmetic---on standardized Sachs data some inputs reach $\approx 8$, where an unclipped $\exp$ would diverge---and they make the fitted mechanism (and hence its analytical Jacobian, \S4.2) piecewise. Their effect on function smoothness and identifiability is a theoretical question for future work.

\subsection{Two-Phase Inference}

Using the EML tree model above, we estimate DAG structure and mechanisms in the following two phases.

\textbf{Phase~1: Mechanism pre-training.}
For all $d(d{-}1)$ pairs $(i, j)$, we train EML tree $T_{j,i}$ via Adam to minimize MSE.
We select the best of 3 random initializations per pair to avoid local optima.
Phase~1 outputs R$^2$ scores (predictive power) and trained EML tree parameters for each pair.

\textbf{Phase~2: Greedy DAG construction.}
Edges are greedily added in decreasing R$^2$ order (maintaining acyclicity), and edges with R$^2$ below a threshold are excluded.
Optionally, BIC backward pruning~\citep{schwarz1978bic} can remove spurious edges using OLS regression for each variable's parent set (effective for dense graphs, but limited for nonlinear mechanisms due to the linear OLS approximation).
Phase~2 outputs DAG $G$, with each edge in $G$ associated with its Phase~1 trained EML tree.

\noindent\fbox{\parbox{0.96\columnwidth}{
\textbf{Algorithm 1:} EML-CD Two-Phase Inference \\[2pt]
\textbf{Input:} Observations $X \in \mathbb{R}^{N \times d}$, depth $D$, threshold $\tau$ \\
\textbf{Output:} DAG $G$, closed-form equations $\{T_{j,i}\}$ per edge \\[3pt]
\textit{// Phase 1: Mechanism pre-training} \\
1. For all $d(d{-}1)$ pairs $(i,j)$: \\
\quad a. Train EML tree $T_{j,i}$ via Adam (MSE), best of 3 random inits \\
\quad b. Compute $R^2_{i \to j}$ \\[3pt]
\textit{// Phase 2: Greedy DAG construction} \\
2. Sort edge candidates by $R^2$ (descending) \\
3. $G \leftarrow$ empty graph \\
4. For each candidate $(i,j)$: if $R^2_{i \to j} > \tau$ and $G + (i{\to}j)$ is acyclic, add to $G$ \\
5. Return $G$ and trained EML trees $\{T_{j,i}\}$
}}

\vspace{4pt}
\textbf{Hyperparameters.}
For Sachs: $D{=}3$, R$^2$ threshold 0.05, 5000 steps, no BIC pruning, learning rate lr=0.01.

\textbf{Design challenges.}
(C1)~Zero initialization of output scale $s_{j,i}$ avoids empty-graph bias.
(C2)~Custom VJP prevents gradient explosion.
(C3)~R$^2$ threshold and optional BIC pruning suppress spurious edges from indirect effects.

\textbf{Direction identification.}
The model $X_j = \sum_i s_{j,i} T_{j,i}(X_i) + \varepsilon_j$ is an additive noise model (ANM)~\citep{hoyer2009anm,peters2014anm}, fit by least squares in Phase~1 (a Gaussian likelihood with fixed variance).
However, EML-CD orients each pair by the \emph{fit-quality} asymmetry of the restricted EML class ($R^2_{i\to j}$ vs.\ $R^2_{j\to i}$, resolved globally by the greedy acyclic ordering), not by a residual-independence test.
This is a heuristic proxy for ANM identifiability: restricting the function class makes the anticausal mechanism harder to represent, lowering its $R^2$.
The principled ANM criterion, however, is that an additive model admits an \emph{independent} residual only in the causal direction~\citep{hoyer2009anm,peters2014anm}---a property that $R^2$ magnitude does not test directly and that can fail when the class is flexible or misspecified.
Replacing the $R^2$ ordering with a residual-independence score (e.g., HSIC, as in RESIT~\citep{peters2014anm}), and replacing the linear OLS used in BIC pruning with the EML mechanism, are important directions for future work.

\subsection{Interpretable Output}

By property~(iii) of \S2.2, EML-CD provides two interpretable outputs for each detected edge:
(a)~a closed-form causal equation directly readable from the EML tree structure, and
(b)~an analytical Jacobian $\partial X_j / \partial X_i$ (causal effect function) derived from that equation.
Edges where the gates along the realized path all bypass degenerate to linear equations, with data-driven selection between linear and nonlinear functions.
These outputs are unavailable from existing methods that output only DAG structure (edge presence) or NN numerical Jacobians.
\S4.2 presents concrete examples from detected Sachs edges.

\section{Experiments}

We verify EML-CD's claim: can structure recovery accuracy be maintained while adding interpretable, closed-form causal mechanisms?
We evaluate three complementary settings: Sachs protein signaling, a synthetic symbolic SEM where true functions are known, and a Causal Chambers light-tunnel subset governed by physical optics.

\subsection{Real Data: Sachs}

Sachs protein signaling data~\citep{sachs2005causal} ($d{=}11$, $N{=}853$, consensus DAG with 17 edges) provides a fair comparison where all methods are evaluated on identical data.
Baselines: PC, GES, CAM (spline additive model), DirectLiNGAM~\citep{shimizu2011directlingam}, NOTEARS, DAGMA, GraNDAG, GOLEM.

\begin{table}[h]
\caption{Sachs protein signaling ($d{=}11$, $N{=}853$). EML-CD: 5 seeds; others: deterministic, single result.}
\centering\small
\begin{tabular}{lcccc}
\toprule
Method & SHD$\downarrow$ & F1$\uparrow$ & Precision & Recall \\
\midrule
\textbf{EML-CD} & $11.2 \pm 0.4$ & $\mathbf{0.492}$ & $\mathbf{0.756}$ & 0.365 \\
PC            & 11 & 0.387 & 0.429 & 0.353 \\
GES           & 11 & 0.387 & 0.429 & 0.353 \\
CAM (spline)  & 12 & 0.483 & 0.583 & 0.412 \\
NOTEARS       & 12 & 0.387 & 0.429 & 0.353 \\
DAGMA         & 13 & 0.370 & 0.500 & 0.294 \\
DirectLiNGAM  & 14 & 0.273 & 0.600 & 0.176 \\
GraNDAG       & 15 & 0.200 & 0.667 & 0.118 \\
GOLEM         & 29 & 0.146 & 0.125 & 0.176 \\
\bottomrule
\end{tabular}
\end{table}

EML-CD achieves SHD=$11.2 \pm 0.4$, on par with PC/GES~(11) and stable across all 5 seeds (SHD $\in [11, 12]$).
Structure recovery on Sachs observational data has been reported to peak at SHD${\approx}12$~\citep{brouillard2024landscape}.
EML-CD's mean SHD (11.2, range $[11,12]$) sits within this reported plateau, on par with PC/GES~(11) and below the closest additive baseline, CAM (SHD=12); the distinguishing feature of EML-CD over CAM is that it outputs \textbf{closed-form equations} rather than non-extractable spline approximations.
Precision 0.756 is the highest among all methods---most detected edges are correct---though recall is only 0.365 (about 6 of 17 true edges detected): EML-CD favors precision over coverage.

\begin{figure}[h]
\centering
\includegraphics[width=\textwidth]{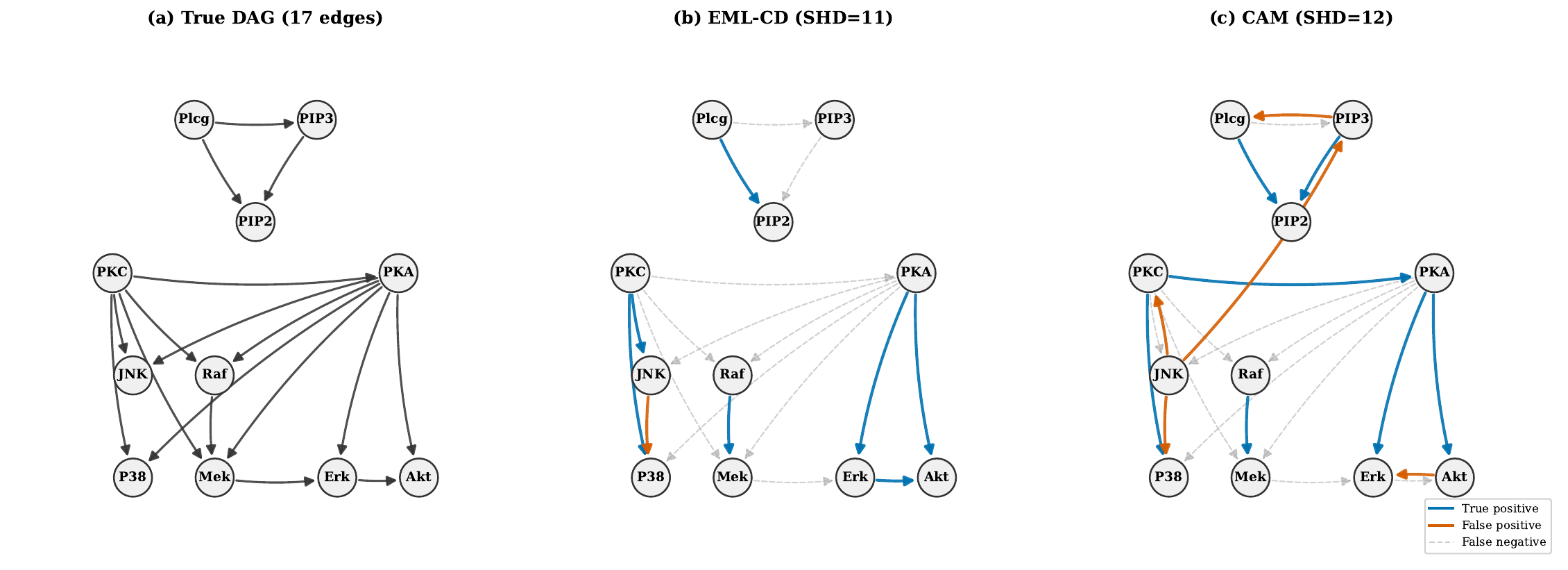}
\caption{True DAG (left), EML-CD (center, representative seed, SHD=11), and CAM (right, SHD=12) on Sachs data. Blue: true positives, orange: false positives, gray dashed: false negatives. EML-CD achieves fewer false positives (1 vs 5) and higher precision than CAM.}
\label{fig:sachs_dag}
\end{figure}

Among the roughly 6 true positive edges on average, known signaling pathways such as Raf $\to$ Mek (MAPK cascade), PKA $\to$ Erk/Akt, and PKC $\to$ P38/JNK are included, providing mechanistic hypotheses directly verifiable by domain experts (Figure~\ref{fig:sachs_dag}).

\subsection{Mechanism Recovery and Causal Effect Functions}

We present concrete examples of the closed-form equations and causal effect functions output by EML-CD for edges detected in \S4.1 (standardized data values).
The core advantage of EML-CD is that it outputs not only DAG structure (edge presence) but also the \textbf{functional form} of each edge in closed form.
This enables (1)~analytical understanding of causal mechanisms, (2)~quantification of state-dependent causal effects via Jacobians, and (3)~automatic discrimination between linear and nonlinear mechanisms.

\begin{figure}[h]
\centering
\includegraphics[width=0.55\textwidth]{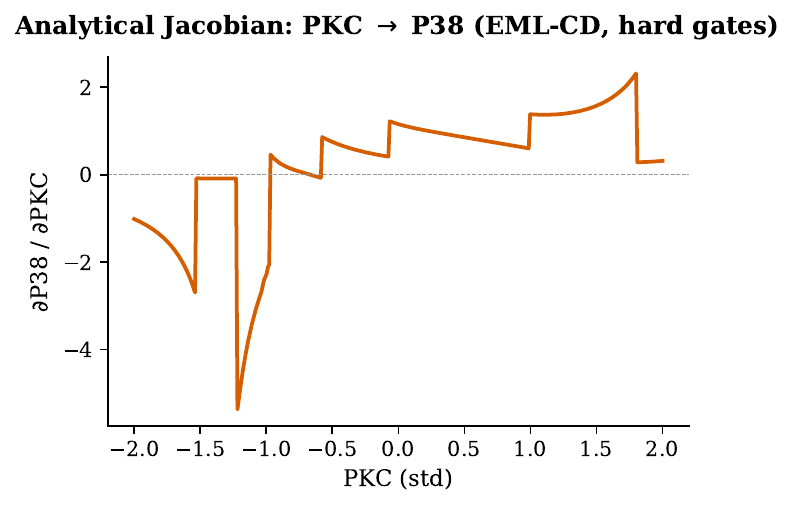}
\caption{Sachs data: analytical Jacobian $\partial$P38/$\partial$PKC for PKC $\to$ P38, computed by automatic differentiation of the gate-annealed (hard-gate) EML-CD mechanism---so the curve is the exact derivative of the displayed Example-1 equation (dense grid; no interpolation). It ranges from $\approx-5.4$ to $\approx2.3$ and changes sign with input level, quantifying a state-dependent nonlinear causal effect; the jumps reflect the numerical bounds of \S3.1. This output is unavailable from methods that return only DAG structure.}
\label{fig:causal_effect}
\end{figure}

\textbf{Example~1: PKC $\to$ P38 (nonlinear mechanism).}
Detected as a depth-3 EML composition (R$^2{=}0.63$). After gate annealing (\S3.1) the gates are hard, so the displayed equation \emph{is} the fitted mechanism:
\begin{equation*}
\resizebox{\columnwidth}{!}{$
\mathrm{P38} = 0.80\,\mathrm{eml}\!\bigl(\mathrm{eml}(0.63\,\mathrm{PKC}{-}1.37,\, \mathrm{eml}(0.31\,\mathrm{PKC}{+}0.43,\, {-}1.96\,\mathrm{PKC}{-}0.65)),\, \mathrm{eml}(\mathrm{eml}(0.07\,\mathrm{PKC}{+}0.34,\, {-}1.77\,\mathrm{PKC}{-}2.22),\, \mathrm{eml}(0.84\,\mathrm{PKC}{-}0.99,\, 0.25\,\mathrm{PKC}{-}1.68))\bigr)
$}
\end{equation*}
The analytical Jacobian $\partial \mathrm{P38} / \partial \mathrm{PKC}$ is obtained by automatic differentiation of this equation (Figure~\ref{fig:causal_effect}; the numerical bounds of \S3.1 make it piecewise).
It ranges from about $-5.4$ (an inhibitory effect at low PKC $\approx -1.2$) to $+2.3$ (a promoting effect at high PKC), changing sign with PKC level.
Such \textbf{state-dependent causal effect} quantification cannot be obtained from DAG structure or linear coefficients alone.
NN-based methods (e.g., GraNDAG) can also compute numerical Jacobians, and \citet{waxman2024dagmadce} show that such derivative-based measures are a valid interpretable signal. What NN models do not expose is a \textbf{closed-form symbolic mechanism}: the underlying function is non-identifiable from the network weights (\S2.1), so only a numerical sensitivity is available.
Given a fitted EML tree, by contrast, the Jacobian is the exact analytical derivative of an \textbf{explicit closed-form equation}. Like any estimator the fitted equation is itself seed-dependent (see below), so EML-CD does not resolve mechanism identifiability; its advantage is the explicit, symbolic, differentiable functional form rather than uniqueness.

\textbf{Example~2: Erk $\to$ Akt (linear mechanism).}
Here every gate along the realized path bypasses, so the mechanism collapses to a linear form (R$^2{=}0.62$):
\[
\mathrm{Akt} = 1.093 \cdot (1.118\,\mathrm{Erk}) \approx 1.222\,\mathrm{Erk}
\]
The Jacobian is constant at $1.222$ regardless of Erk level, indicating a linear causal effect.
Because the gates are hard after annealing, this linear form is the exact fitted mechanism (faithful, R$^2{=}0.62$): the depth gate $\sigma(\gamma)$ performs data-driven selection between linear and nonlinear functions (\S3.1) without adding unnecessary nonlinear terms.

\textbf{Example~3: Raf $\to$ Mek (nonlinear mechanism).}
A key edge in the MAPK cascade, detected with the highest R$^2{=}0.64$ among true positive edges:
{\small
\[
\mathrm{Mek} = 1.41 \cdot \mathrm{eml}\!\bigl(0.35\,\mathrm{Raf} - 0.50,\; \mathrm{eml}({-}0.21\,\mathrm{Raf} + 0.45,\; {-}0.05\,\mathrm{Raf} + 0.27)\bigr)
\]
}
Its analytical Jacobian is positive throughout ($\partial\mathrm{Mek}/\partial\mathrm{Raf} \in [0.37, 0.78]$), i.e.\ a monotone activating effect consistent with the well-established Raf $\to$ Mek $\to$ Erk cascade.

\textbf{Recovery accuracy.}
The R$^2$ values of detected true-positive edges range from about 0.07 to 0.64, with the strongest (Raf$\to$Mek, PKC$\to$P38, Erk$\to$Akt) near 0.6, i.e.\ each such EML tree explains roughly half of the target variable's variance. The residual is attributable to additive noise and undetected parent variables (false negative edges in \S4.1, e.g., PKA $\to$ P38).
The output equations are thus \textbf{partial approximations} of the causal mechanisms; higher R$^2$ is expected when all parent variables are detected.
Furthermore, since the method depends on random initialization, different seeds can yield equations with different parameters for the same edge. Improving output equation stability is a direction for future work.

\subsection{Controlled Mechanism Recovery}

\begin{figure}[h]
\centering
\includegraphics[width=\textwidth]{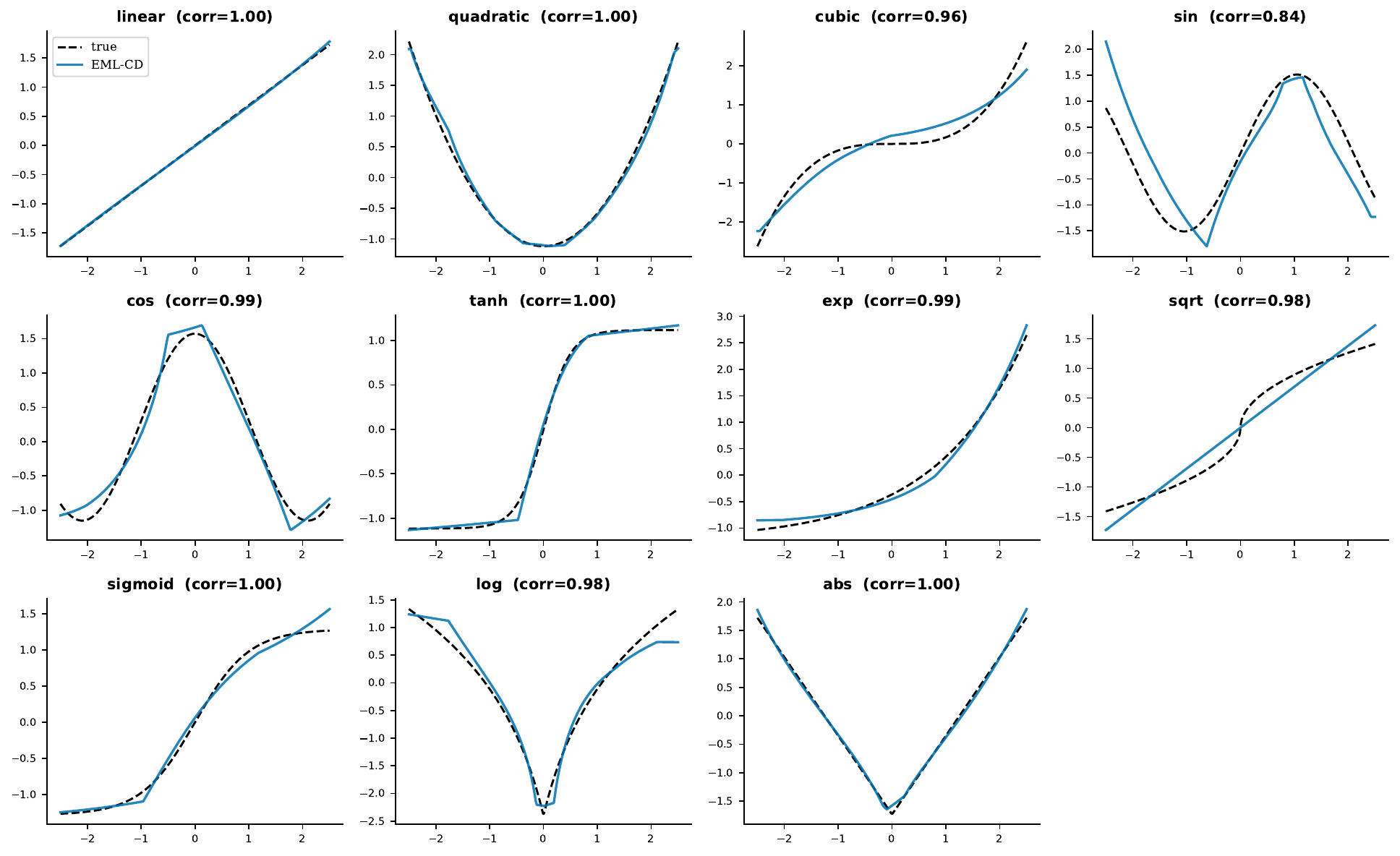}
\caption{Controlled mechanism recovery: true mechanism (dashed) vs.\ the recovered gate-annealed EML-CD closed form (blue), standardized, on a held-out grid. Ten of eleven elementary families are recovered faithfully (shape correlation in each panel title); only high-frequency $\sin$ (top right) is partial.}
\label{fig:mech_recovery}
\end{figure}

To test whether EML-CD recovers the \emph{true} mechanism---not merely emits an expression---we isolate mechanism recovery from structure learning. For a bivariate $Y = f(X) + \varepsilon$ with known $f$ (high SNR, $N{=}2000$, $X$ uniform on $[-2.5,2.5]$), we fit a depth-3 gate-annealed EML tree and compare the recovered closed form $\hat f$ to $f$ on held-out data across eleven elementary families. Because the gates are hard after annealing (\S3.1), $\hat f$ is exactly the displayed equation.
EML-CD recovers ten of eleven families with held-out shape correlation $\geq 0.96$---linear, quadratic, cubic, $\cos$, $\tanh$, $\exp$, $\sqrt{\cdot}$, sigmoid, $\log$, and $|x|$ (Figure~\ref{fig:mech_recovery}); only the high-frequency $\sin(1.5x)$ is partial (0.84).
This confirms that the closed-form output is a faithful recovery of the underlying mechanism, not merely a fitted curve. It also explains the weaker embedded result in \S4.4: the families that degrade there ($\tanh$, $\cos$) are recovered cleanly here, so that degradation reflects depth-2 trees, lower SNR, and interaction with structure learning rather than a representational limit of EML trees. This is a favorable setting (isolated edge, high SNR, multiple restarts); recovery on noisy, embedded data is harder.

\subsection{Synthetic Symbolic Mechanism Recovery}

Sachs evaluates scientific plausibility but does not provide ground-truth functional mechanisms.
To directly test mechanism recovery, we construct S-Sym, a symbolic SEM benchmark with known DAG and known edge functions.
Graphs are ER DAGs with $d{=}10$, $N{=}2000$, and average 1.5 parents per node.
Each true edge is assigned one of five mechanisms well covered by elementary functions: linear, $\sin$, quadratic, $\cos$, or $\tanh$, with random scale.
We evaluate both structure recovery and function recovery on held-out samples.

For a true edge $(i,j)$, function MSE is
\[
  \mathrm{fMSE}_{i \to j}
  = \frac{1}{N_{\mathrm{ho}}}\sum_{n=1}^{N_{\mathrm{ho}}}
      \left(f_{i \to j}^{\star}(x_i^{(n)}) -
            \hat f_{i \to j}(x_i^{(n)})\right)^2 .
\]
For the SINDy dictionary baseline, we also report mechanism recovery score (MRS), the fraction of correctly detected true edges whose dominant basis matches the true function family.

\begin{table}[h]
\caption{S-Sym symbolic SEM benchmark ($d{=}10$, $N{=}2000$, 3 seeds). f-MSE is computed on held-out observations for true edges.}
\centering\small
\begin{tabular}{lcccc}
\toprule
Method & SHD$\downarrow$ & F1$\uparrow$ & MRS$\uparrow$ & f-MSE$\downarrow$ \\
\midrule
SINDy dictionary (5 bases) & $14.3 \pm 1.2$ & 0.325 & $0.257 \pm 0.239$ & 7644.46 \\
\textbf{EML-CD} & $14.0 \pm 3.3$ & \textbf{0.501} & -- & \textbf{3.67} \\
\bottomrule
\end{tabular}
\end{table}

\begin{figure}[h]
\centering
\includegraphics[width=0.65\textwidth]{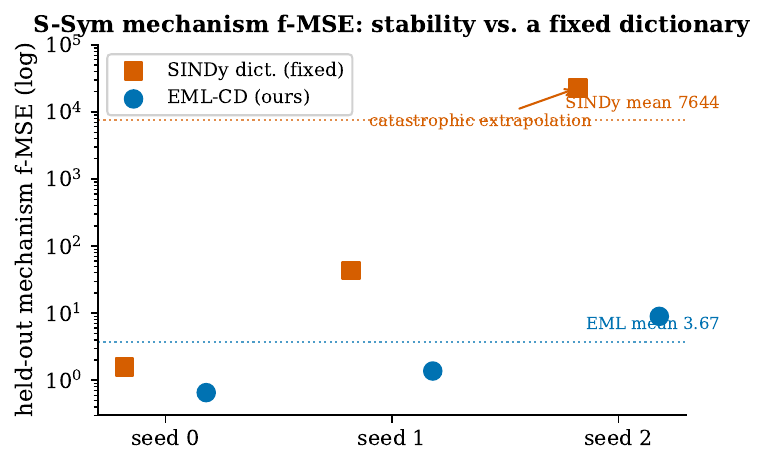}
\caption{S-Sym held-out mechanism f-MSE per seed (log scale). EML-CD (blue) stays low and stable across seeds, whereas the fixed SINDy dictionary (orange) extrapolates catastrophically on one seed ($\approx$2.3$\times10^4$), driving its high mean (7644 vs.\ 3.67). The benefit is f-MSE \emph{stability}, not faithful recovery of every functional shape (see text).}
\label{fig:ssym_mechanism}
\end{figure}

The structural SHD is still not competitive with specialized linear optimizers, and the mechanism result is best read as a \emph{stability} result rather than faithful shape recovery: the fixed SINDy dictionary is brittle---one seed yields a catastrophic held-out f-MSE ($\approx$2.3$\times10^4$) because the selected basis extrapolates poorly---whereas EML-CD stays low and stable across seeds despite using the same greedy structural selection (Figure~\ref{fig:ssym_mechanism}).
At the level of individual mechanisms this stability is uneven: near-linear edges are recovered accurately, but at depth~2 and this sample size the EML tree only partially captures strong nonlinearities (e.g., the saturation of $\tanh$) and does not reliably recover $\cos$. Faithful shape recovery for highly nonlinear families is left for future work.

\subsection{Causal Chambers Light-Tunnel Subset}

We next evaluate practical usefulness on Causal Chambers~\citep{gamella2025causalchambers}, a real physical benchmark with known causal ground truth and interventional data.
We use the light-tunnel Malus-law subset ($d{=}6$), where optical intensity is governed by interactions such as $\cos^2(\theta_1-\theta_2)$.
This is a deliberately challenging setting for the additive edge model in \S3.1 because true physics contains interactions between variables.

\begin{table}[h]
\caption{Causal Chambers light-tunnel Malus subset ($d{=}6$).}
\centering\small
\begin{tabular}{lcccc}
\toprule
Method & SHD$\downarrow$ & F1$\uparrow$ & Precision & Recall \\
\midrule
Correlation threshold & 7 & 0.000 & 0.000 & 0.000 \\
OLS-BIC & 8 & 0.273 & 0.188 & \textbf{0.500} \\
EML-CD (depth 2) & \textbf{5} & 0.444 & \textbf{0.667} & 0.333 \\
EML-CD (depth 3) & 7 & 0.000 & 0.000 & 0.000 \\
Interaction-EML (pair leaves) & 6 & \textbf{0.500} & 0.500 & \textbf{0.500} \\
\textbf{Interaction-EML + intervention pruning} & \textbf{4} & \textbf{0.500} & \textbf{1.000} & 0.333 \\
\bottomrule
\end{tabular}
\end{table}

Depth-2 EML-CD improves over linear OLS-BIC in SHD, F1, and precision, indicating that compact EML trees can capture useful nonlinear signal in a physical system.
Depth-3 fails in this small sample setting, showing that additional tree capacity is not automatically beneficial.
As a first interaction-aware test, we also replace univariate leaves $a_l x_i + b_l$ with pairwise affine leaves $a_l^\top X_S + b_l$ and score singleton/pair parent sets.
Without causal pruning this improves F1 from 0.444 to 0.500 by increasing recall, but it also introduces descendant/correlation false positives.
Using the intervention-file metadata to restrict parent candidates to directly intervened variables removes these false positives, yielding SHD=4 and precision=1.000, although recall remains 0.333.
The next stage is therefore not only to represent the child mechanism as $T_j(X_{\mathrm{pa}(j)})$ for terms like $\cos^2(\theta_1-\theta_2)$, but also to combine multivariate mechanism fitting with interventional context or conditional-independence constraints for parent-set pruning.

\section{Related Work}

We clarify EML-CD's positioning relative to three related research areas.

\textbf{Structure learning.}
NOTEARS~\citep{zheng2018notears}, GOLEM~\citep{ng2020golem}, DAGMA~\citep{bello2022dagma} (linear); GraNDAG~\citep{lachapelle2020grandag}, NOTEARS-MLP~\citep{zheng2020notears_mlp} (nonlinear). All produce point estimates without mechanism output. DiBS~\citep{lorch2021dibs} provides Bayesian inference over DAG posteriors but uses NN likelihoods, leaving mechanisms unidentified. EML-CD's initial implementation is based on the DiBS codebase, but replaces Bayesian DAG posterior inference with pairwise pre-training and greedy DAG construction.
A recent differentiable constraint-based method~\citep{zhou2025dagpa} improves conditional-independence testing via gradient optimization, but still targets graph recovery rather than closed-form mechanism recovery.

\textbf{Interpretable causal models.}
CAM~\citep{buhlmann2014cam} uses additive splines but does not output symbolic equations (SHD=12 on Sachs, \S4.1). BF-BIC~\citep{ramsey2025bfbic} uses Legendre bases for DAG scoring but does not estimate functions. DAGMA-DCE~\citep{waxman2024dagmadce} provides numerical Jacobians but not closed-form equations. EML-CD differs from all of these by outputting both closed-form equations and analytical Jacobians.
MDL-based causal direction methods~\citep{brogueira2026rdmdl} are closely related in spirit: they select causal direction by favoring simpler mechanisms. EML-CD can be viewed as an operational, multivariate approximation of this principle because tree depth and gate activation define an explicit description length proxy.

\textbf{Symbolic regression.}
SINDy~\citep{brunton2016sindy} inspired symbolic basis dictionaries, and the EML operator~\citep{odrzywolk2026eml} introduced a single binary operator whose gated trees, trainable by gradient descent, compose elementary functions; we adopt this learnable-circuit formulation as our mechanism representation. Symbolic equations have also been extracted post-hoc from trained models---e.g.\ from GNN edge functions on a \emph{given} graph~\citep{cranmer2020symbolic}---and KaCGM~\citep{almodovar2026kacgm} learns KAN (Kolmogorov-Arnold Network) mechanisms on a \emph{known} DAG and extracts symbolic equations post-hoc; neither performs DAG structure learning. To our knowledge, EML-CD is the first method to integrate gradient-based symbolic regression \emph{into} causal structure learning so that closed-form equations are attached to each edge of a DAG discovered from observational data. We note that the EML operator is itself very recent and not yet independently validated at scale, which we regard as a limitation of the present foundation.

\section{Discussion and Limitations}

\textbf{Model assumptions.}
The method relies on two assumptions: (1)~the univariate-additive model $X_j = \sum_i s_{j,i} T_{j,i}(X_i) + \varepsilon$ cannot represent cross-variable interactions (e.g., $\cos^2(\theta_1 - \theta_2)$); the Interaction-EML variant (\S4.5) partially relaxes this through pairwise-affine leaves $a_\ell^\top X_S + b_\ell$, which give the tree access to combinations such as $\theta_1-\theta_2$, but it still does not provide exact multivariate mechanisms $T_j(X_{\mathrm{pa}(j)})$; and (2)~causal sufficiency (no unobserved confounders). Extending to fully interaction-aware mechanisms and latent-variable methods such as FCI are directions for future work.
The Causal Chambers result makes the first limitation concrete: depth-2 trees recover some physical edges, pairwise interaction leaves improve recall, and intervention-based parent pruning improves precision.
However, recall remains low for the optical intensity nodes, so the interaction extension must be evaluated as a causal parent-selection problem, not merely as a regression improvement.

\textbf{Evaluation scope and scalability.}
This study evaluates Sachs, S-Sym, and a Causal Chambers subset; the full ER/SF nonlinear benchmark grid used by DAGMA-DCE and recent differentiable constraint-based experiments is not yet covered.
Depth $D$ and the R$^2$ threshold were tuned per setting.
Pre-training scales as $O(d^2)$ pairs; $d{=}10$ takes approximately 85 seconds (CPU), with $d \leq 20$ as the practical upper limit.

\subsection{Future Work}
The present study establishes EML-CD as a proof of concept for interpretable mechanism recovery; several directions remain open and are explicitly left for future work (none of the following were evaluated here).
(1)~\textbf{Full nonlinear benchmark grid.} We plan a systematic evaluation on the Erd\H{o}s--R\'enyi (ER) and scale-free (SF) nonlinear benchmark grid used by DAGMA-DCE~\citep{waxman2024dagmadce} and recent differentiable constraint-based experiments~\citep{zhou2025dagpa}, sweeping graph density and node count, to characterize where EML trees help and where structural recovery degrades.
(2)~\textbf{Automatic hyperparameter selection.} Depth $D$ and the $R^2$ edge threshold $\tau$ are currently tuned per setting; we will develop data-driven selection (e.g., held-out $R^2$ or description-length criteria) so that complexity is chosen without manual tuning.
(3)~\textbf{Multivariate, interaction-aware mechanisms.} Extending edge functions from univariate $T_{j,i}(X_i)$ to multivariate $T_j(X_{\mathrm{pa}(j)})$ (e.g., to represent $\cos^2(\theta_1-\theta_2)$) must be formulated as a causal \emph{parent-selection} problem, combining multivariate mechanism fitting with interventional context or conditional-independence constraints, rather than treated as a pure regression-accuracy improvement.
(4)~\textbf{Bayesian SCM extension.} Fixing the closed-form equations output by EML-CD as structural causal model (SCM) equations and performing Bayesian posterior inference over their parameters would jointly deliver interpretable equations and uncertainty quantification, integrating the Bayesian DAG inference of DiBS~\citep{lorch2021dibs} with EML-tree interpretability.
(5)~\textbf{Scalability beyond $d\leq 20$.} Pre-training scales as $O(d^2)$ pairs; scaling to larger graphs will require candidate-pair pruning, parallelization, or amortized pre-training.
(6)~\textbf{Output-equation stability.} Because pre-training depends on random initialization, different seeds can yield different parameterizations for the same edge; improving the stability and identifiability of the recovered closed-form equations across seeds is needed.
(7)~\textbf{Identifiability theory under gradient clipping.} The custom VJP clips node gradients to $[-10,10]$ and floors the $\ln$ argument at $0.5$; the effect of this clipping on function smoothness and on mechanism identifiability remains an open theoretical question.

\section{Conclusion}

EML-CD integrates EML symbolic trees~\citep{odrzywolk2026eml} into causal structure learning, targeting interpretable mechanism recovery.
It circumvents the ill-posedness of NN-ANM mechanism extraction (\S2) through finite-parameter EML function representation, attaching closed-form equations and analytical Jacobians to each edge.
On real data (Sachs, SHD=$11.2 \pm 0.4$ over 5 seeds, precision 0.756), structure-recovery accuracy is on par with PC/GES within seed variance, while closed-form equations and analytical causal-effect functions are attached to each edge.
On a symbolic synthetic benchmark the EML tree attains a far lower and more stable mechanism f-MSE than a fixed dictionary (3.67 vs.\ 7644), though its structural accuracy only matches the dictionary baseline and its shape recovery is reliable mainly for near-linear mechanisms; on a Causal Chambers subset a depth-2 model beats linear OLS-BIC in F1, but the gain is configuration-dependent (depth-3 collapses) and the strongest precision relies on intervention metadata.
On the detected Sachs edges, depth gates select between linear and nonlinear forms in a data-driven manner, illustrated qualitatively by the recovered equations; a quantitative evaluation of selection correctness is left for future work.
This work provides initial evidence that symbolic causal-mechanism interpretability can be added without sacrificing structure-recovery accuracy.

\bibliographystyle{plainnat}
\bibliography{references}

\end{document}